\documentclass{bmvc2k}
\usepackage{booktabs}
\usepackage{amssymb}
\usepackage{mlmath}
\usepackage{wrapfig}
\usepackage{multirow}
\usepackage[nameinlink]{cleveref}
\crefname{figure}{}{}
\newcommand{\paren}[1]{\mathopen{}\left( {#1}_{{}_{}}\,\negthickspace\right)\mathclose{}}

\title{Dual-Query Multiple Instance Learning for Dynamic Meta-Embedding based Tumor Classification}

\addauthor{Simon Holdenried-Krafft}{simon.krafft@uni-tuebingen.de}{1}
\addauthor{Peter Somers}{somers@isys.uni-stuttgart.de}{3}
\addauthor{Ivonne A. Montes-Majarro}{ivonne.montes@med.uni-tuebingen.de}{2}
\addauthor{Diana Silimon}{dianasilimon@icloud.de}{2}
\addauthor{Cristina Tarín}{cristina.tarin-sauer@isys.uni-stuttgart.de}{3}
\addauthor{Falko Fend}{falko.fend@med.uni-tuebingen.de}{2}
\addauthor{Hendrik P. A. Lensch}{hendrik.lensch@uni-tuebingen.de}{1}

\addinstitution{
 Institute for Computer Graphics,\\
 University of Tübingen,\\
 Tübingen, Germany
}
\addinstitution{
Institute of Pathology and Neuropathology,\\
University Hospital of Tübingen,\\
Tübingen, Germany
}
\addinstitution{
Institute for System Dynamics,\\
University of Stuttgart,\\
Stuttgart, Germany
}

\runninghead{S.Holdenried-Krafft et al.}{Dual-Query Multiple Instance Learning}

\begin{document}

\maketitle

\begin{abstract}
Whole slide image (WSI) assessment is a challenging and crucial step in cancer diagnosis and treatment planning. WSIs require high magnifications to facilitate sub-cellular analysis. Precise annotations for patch- or even pixel-level classifications in the context of gigapixel WSIs are tedious to acquire and require domain experts. Coarse-grained labels, on the other hand, are easily accessible, which makes WSI classification an ideal use case for multiple instance learning (MIL). In our work, we propose a novel embedding-based Dual-Query MIL pipeline (DQ-MIL). We contribute to both the embedding and aggregation steps. Since all-purpose visual feature representations are not yet available, embedding models are currently limited in terms of generalizability. With our work, we explore the potential of dynamic meta-embedding based on cutting-edge self-supervised pre-trained models in the context of MIL. Moreover, we propose a new MIL architecture capable of combining MIL-attention with correlated self-attention. The Dual-Query Perceiver design of our approach allows us to leverage the concept of self-distillation and to combine the advantages of a small model in the context of a low data regime with the rich feature representation of a larger model. We demonstrate the superior performance of our approach on three histopathological datasets, where we show improvement of up to 10\% over state-of-the-art approaches. GitHub repository: \url{https://github.com/cgtuebingen/DualQueryMIL} 
\end{abstract}

\section{Introduction}
\label{sec:intro}
Histopathological slide assessment is the gold standard for grading and treatment planning for almost all types of cancer \cite{Gurcan2009}. In computational pathology, slide scanners convert tissue specimens on glass slides into digital images. Due to the required subcellular details, the scanned slide specimens, also called whole slide images (WSIs), can have more than a hundred thousand pixels in each dimension. Processing such gigapixel images entirely is computationally intractable. Hence, WSIs are subdivided into patches, reducing the computational burden and allow for  processing each patch with well-established architectures such as a convolutional neural network (CNN) or Transformers \cite{Vaswani2017}. Unfortunately, precise annotations for patch- or even pixel-level classifications in the context of gigapixel images are labor intensive to acquire and require expert knowledge. Instead, slide-level labels, such as tissue type, cancer grade, or molecular subtype are widely available and less time-consuming to collect. Multiple instance learning (MIL), a subset of weakly supervised learning introduced by \citet{Dietterich1997}, can make use of such coarse-grained labels and has shown its effectiveness in the field of WSI classification in a variety of recent studies \cite{Ilse2018,Li2020,Campanella2019,Hou2015,Lu2020,Wu2022}.

MIL defines one sample as a bag of instances and there are two major categories: instance-based or embedding-based~\cite{Amores2013, Ilse2018}. Different studies indicate that embedding-based MIL has superior performance compared to instance-based MIL~\cite{Campanella2019,Ilse2018,Li2020,Qian2022,Shao2021}. Embedding-based approaches first transform all instances into learned feature vectors, aggregate them into a joint bag representation, and conclude with a bag-level classification.
The initial step of acquiring robust visual features is demanding, especially for WSI classification, where relevant features depend on the cancer entity~\cite{Stacke2022}. But even for the same entity, WSIs can vary from hospital to hospital and show severe differences in appearance due to slightly different staining chemicals \cite{Sikaroudi2023}. Thus, out-of-distribution generalization remains a challenge for embedding models, and as the quality of the feature representations directly affects the performance on the downstream task \cite{Li2020}, it is not negligible.

Although aggregation models can supplement the embedding architecture by leveraging the supervised training signal \cite{Kiela2018} to enrich the representations, they come with inherent issues. In classical MIL, a WSI is defined as a bag and its corresponding patches are assumed to be independent and identically distributed (i.i.d.) instances. Given a binary cancer classification task, the whole bag is labeled as cancerous as soon as a single patch is cancerous. For highly unbalanced bags, where only a small fraction of patches are actually positive (diseased), the training signal diminishes due to the dominance of negative instances \cite{Li2020,Zhang2022}. In such cases, simple models tend to misclassify. While larger models can still learn rich bag representations, they tend to overfit in small data regimes, common in medical image analysis. Another dubious aspect of classical MIL in the context of WSI classification is the i.i.d.~assumption \cite{Shao2021,Tu2019}. In fact, pathologists exploit structural information to enrich smaller areas with the surrounding context. Thus, correlating instances seems natural, but due to a large number of instances within one bag, it can be computationally demanding, especially for Transfomer-based approaches \cite{Shao2021}.

In our work, we address the various topics previously mentioned. We conduct extensive experiments to validate the benefits of our approach and evaluate our method based on three different publicly available medical datasets on tumor classification and cancer subtyping. Our contributions are threefold:
\begin{itemize}
	 \item We introduce a novel MIL architecture, named Dual-Query MIL inspired by the Perceiver \cite{Jaegle2021}. Due to its design, the Perceiver decouples the input size from the dimensionality of a latent representation, eliminating the quadratic scaling problem of the classical Transformer architecture \cite{Vaswani2017}. Our dual-query design in the cross-attention layer combines i.i.d.~MIL attention \cite{Ilse2018} with correlative self-attention \cite{Vaswani2017} in one architecture.
	 \item We introduce a self-distillation loss function, which allows us to leverage both the advantages of a small and a larger aggregation model in one framework, preventing overfitting while simultaneously acquiring rich feature representations.
	 \item We explore the potential of dynamic meta-embedding \cite{Kiela2018, Truong2021} based on three state-of-the-art self-supervised learning (SSL) methods in the context of MIL. Our experiments show the superiority of dynamic meta-embedding compared to individual embeddings and indicate a step towards robust visual representations in the context of medical image analysis.
\end{itemize}

\section{Related Work}
\label{sec:related_work}
As our work focuses on deep MIL-based histopathological slide assessment, we want to provide an overview of the most recent trends and the role of SSL embedding specific to this field of research. For further literature, we refer to \cite{Amores2013,Carbonneau2018,Wang2022}.

Deterministic MIL pooling operations, such as max or mean pooling, are limited in terms of performance. Therefore, \citet{Ilse2018} base the pooling operation on DNNs, which assign attention scores to i.i.d.~instances, defining the contribution of each instance to the final bag representation. \citet{Lu2020} extended the idea of attention-based instance scoring. They utilize instance-level clustering to guide the learning and to constrain the feature space by creating class-specific pseudo-labels and subsequent class-specific attention branches for multi-class settings. 
All these methods neglect correlations between instances, whereas graph or capsule-based architectures \cite{Yan2018,Tu2019} incorporate contextual information between instances. This resembles a pathologist's proceeding that connects local characteristics such as the nucleus shape and size with the global context, e.g. surrounding cell architecture. Most recent architectures use non-local attention. Dual-stream MIL (DS MIL) \cite{Li2020} consists of one branch, which detects the most significant instance using a max-pooling operation, and a second branch correlating the detected characteristic instance with all remaining instances using a Transformer-like one-to-all attention mechanism. The one-to-many approach by \citet{Bergner2023} similarly consists of two stages: an iterative patch selection (IPS) and a small cross-attention Transformer stage. Using the IPS module drastically reduces the number of patches per bag, accelerating the aggregation step. \citet{Shao2021} utilize an approximated all-to-all self-attention by utilizing the Nyström method \cite{Xiong2021}. This allows for large inputs, as is crucial for WSI classification, and reduces the computational cost of multi-head self-attention.

Our proposed method is based on the Perceiver model by \citet{Jaegle2021}. Instead of the classical all-to-all Transformer self-attention \cite{Vaswani2017} with its quadratic scaling problem, the Perceiver relies on an asymmetric attention mechanism. This reduces the computational complexity and decouples the input size from the depth of the architecture. The Perceiver exploits a cross-attention layer to transform the input into a condensed latent array. The all-to-all self-attention is only applied in this latent array. Furthermore, we combine this approach with the one-to-all query design from \citet{Bergner2023}. Our adaptation combines MIL and Transformer attention in one architecture. Whereas most other methods rely on cross-entropy (CE) loss with bag-labels as the training signal \cite{Ilse2018,Li2020,Shao2021}, we utilize the concept of self-distillation \cite{Zhang2019_SD, Zhang2022_SD} to fully exploit the potential of our approach. Besides the aggregation model, we also explore new ways of feature extraction or merging. As indicated by \citet{Tendle2021}, the generalization of SSL representations is improved compared to supervised learning (SL) representations. However, instead of training an embedding model with histopathological samples using SSL \cite{Li2020, Chen_2022_CVPR}, we explore the potential of dynamic meta-embedding in the context of MIL based on three of the most recent pre-trained SSL methods (SwAV \cite{Caron2020}, DINO \cite{Caron2021}, DINOv2 \cite{Oquab2023}). This idea from the vibrant field of natural language processing showed increased robustness and generalization by combining multiple embedding techniques complementary to one another \cite{Kiela2018, Truong2021}. 
\section{Methodology}
\label{sec:Method}
During classical supervised training, a model learns to estimate the given label $y$ corresponding to input image $x \in \mathbb{R}^{h\times w \times 3}$. Instead, multiple-instance learning is set-based. Each set consists of multiple inputs, or instances, and is called a bag $\fB = \{x_1, ..., x_N\}$. The number of instances $N$ within the bag can vary between bags. Moreover, we assume that there exists a label $y_n$ with $n=1, ..., N$ for each instance within the bag, which is unknown. Only one global label $\fY$ is given for the whole bag $\fB$. In a binary MIL classification task, the bag label is positive as soon as a single instance label is positive. To estimate the final label of bag $\fB$, multiple instance learning requires suitable transformations represented by $f$ and $g$. The choice of $f$ and $g$ defines whether it is an instance-based or embedding-based approach \cite{Amores2013,Ilse2018}. In instance-based MIL, $f$ transforms each instance into scores, and function $g$ is a pooling operation, such as max- or mean-pooling, aggregating the scores. In embedding-based MIL, $f$ first projects the instances into a newly learned embedding space, and function $g$ afterward distills all instances corresponding to one bag into a joint bag representation.

\begin{figure}[!h]
	\centering
	\includegraphics[width=0.9\textwidth]{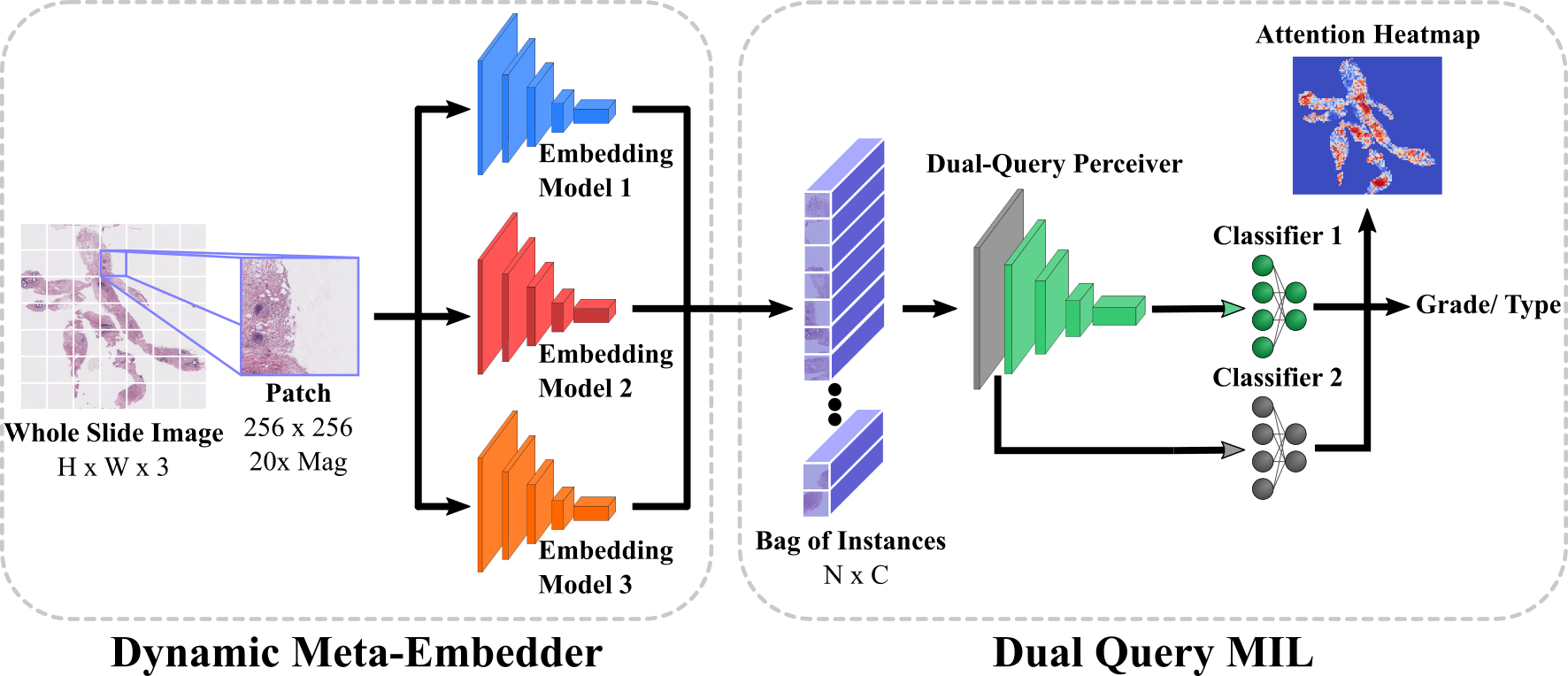}
	\caption{DQ-MIL pipeline. The Dynamic Meta-Embedder (DME) combines three different feature representations per patch and creates one joint feature vector. The bag of DME representations is then processed by the Dual-Query (DQ) Perceiver in two different pathways, exploiting the advantages of MIL- and self-attention.}
	\label{fig:dqmil_pipeline}
\end{figure}

In our proposed method, illustrated in Figure~\ref{fig:dqmil_pipeline}, we touch upon both transformations $f$ and $g$ of the embedding-based procedure. First, we introduce the concept of meta-embedding in the context of multiple instance learning with the Dynamic Meta-Embedder (DME), corresponding to projection $f$. Furthermore, we propose the Dual-Query (DQ) Perceiver representing function $g$, based on the Perceiver architecture \cite{Jaegle2021}. Our method leverages the flexibility of the Perceiver and joins MIL and self-attention in one framework.

\subsection{Dynamic Meta-Embedding for Multiple Instance Learning}
Instance-embedding models transform a raw input patch $x_i$ into a feature vector $\mathbf{h}_i=f(x_i)$. We utilize three SSL pre-trained encoding models to distill the raw image into a single feature representation. Rather than just concatenate the embeddings, we utilize the training signal of the downstream task to dynamically learn the new representation \cite{Kiela2018,Truong2021}. Our Dynamic Meta-Embedder consists of two ResNet50 architectures \cite{He2016}, and one Vision Transformer \cite{Dosovitskiy2020} (ViT-L/14). The two ResNet models were pre-trained on the ImageNet dataset \cite{Deng2009}, whereas the ViT used the LVD-142M dataset \cite{Oquab2023}.

\label{subsec:ssl_meta}
\begin{wrapfigure}{r}{0.45\textwidth}
	\vspace{-6pt}
	\centering
	\includegraphics[width=0.95\linewidth]{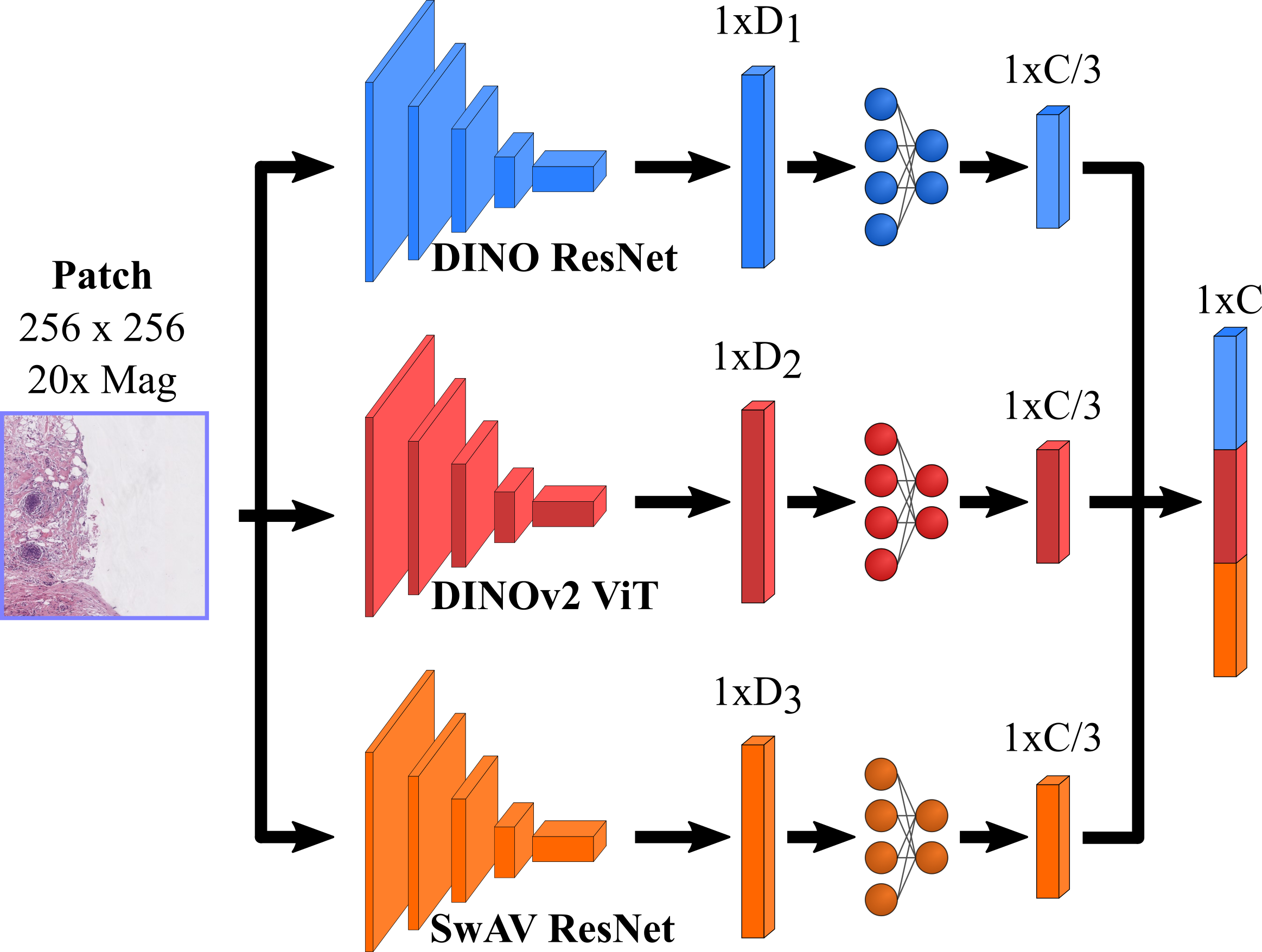}
	\caption{Dynamic Meta-Embedder. The different embedding models first condense each patch into a representation vector, then each of them gets processed in three different trainable linear layers and concatenated to a single vector. }
	\vspace{1pt}
	\label{fig:metaembedding}
\end{wrapfigure}
Besides the architecture, all three embedders differ in terms of the SSL technique used for pretraining. One ResNet model was pre-trained using SwAV \cite{Caron2020}, the other one utilizes the DINO approach \cite{Caron2021}. The ViT model was pre-trained with the most recent SSL method DINOv2 \cite{Oquab2023}. DINOv2 joins ideas from various SSL methods, the image-level loss of DINO \cite{Caron2021}, the masked image modeling of iBOT \cite{Zhou2022}, the Sinkhorn-Knopp centering of SwAV \cite{Caron2020}, and more. After piping the input patch through each of the embedding models, the Dynamic Meta-Embedder projects all three embeddings of various lengths to the same dimensionality using separate linear layers per embedder. This step, in which the three SSL models are frozen, allows exploiting the training signal from the bag label to finetune the representations and to extract task- and domain-specific features. Figure \ref{fig:metaembedding} depicts the DME module.

\subsection{Dual-Query Perceiver}
\label{subsec:dq_perceiver}
\begin{figure}[!h]
	\centering
     {
    \phantomsubcaption\label{fig:dq_perceiver_a}
    \phantomsubcaption\label{fig:dq_perceiver_b}
    }
	\includegraphics[width=1.\textwidth]{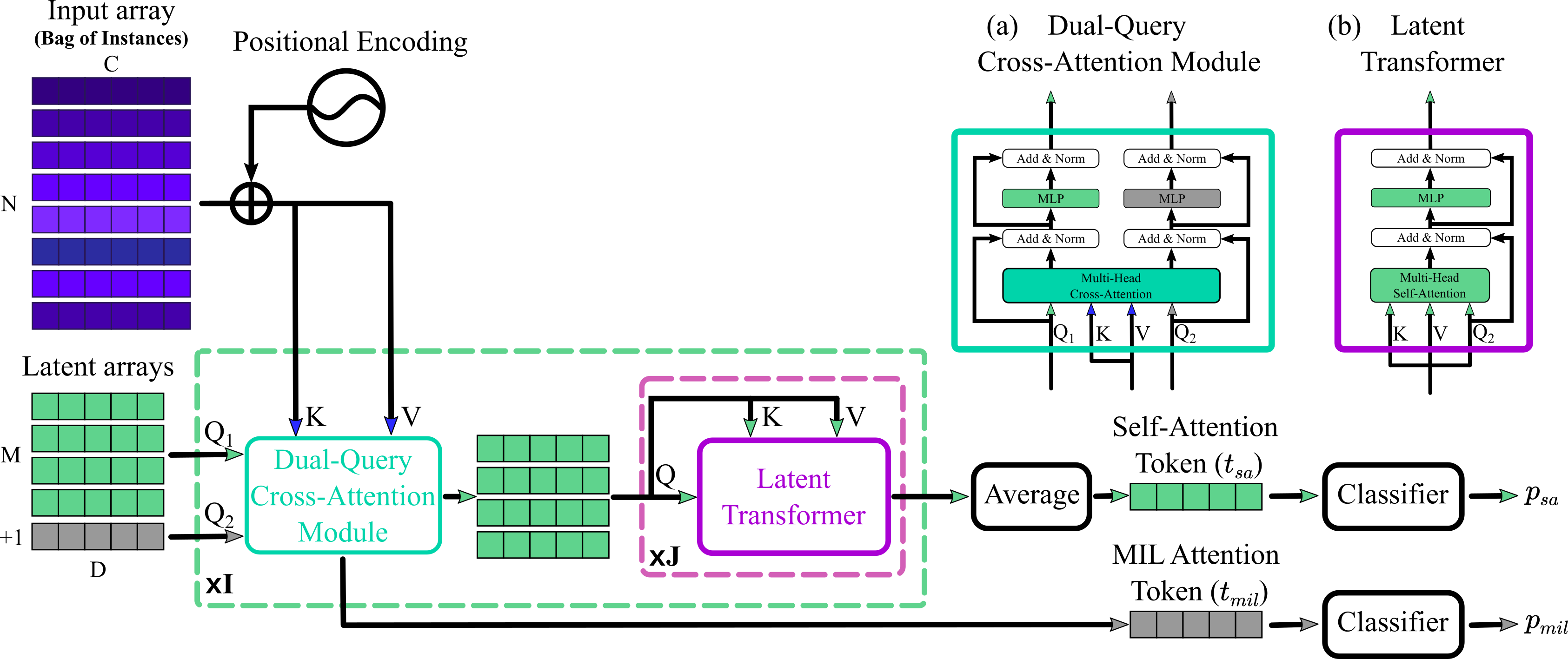}
	\caption{Dual-Query Perceiver. Consisting of two pathways, the DQ Perceiver combines MIL- and self-attention in one framework. The two main components, Dual-Query Cross-Attention Module \textbf{(a)}, and Latent Transformer \textbf{(b)}, follow the typical structure proposed by \citet{Vaswani2017}. However, the proposed Cross-Attention module consists of two pathways based on two separate queries $\mathbf{Q_1}$ and $\mathbf{Q_2}$. Both branches share keys $\mathbf{K}$ and values $\mathbf{V}$.}
	\label{fig:dq_perceiver}
\end{figure}
As an aggregator model to summarize the bag, we propose the Dual-Query Perceiver. This architecture is based on the Perceiver and Perciever IO idea \cite{Jaegle2021, Jaegle2022}. We leverage the flexibility of the proposed querying mechanism and propose the novel MIL architecture in Figure \ref{fig:dq_perceiver}.
The key components of our method are the Dual-Query Cross-Attention Module and the Latent Transformer. Both include a query-key-value (QKV) attention block, the core element in all Transformer-like architectures. It transforms the input into queries $\mathbf{Q}$, keys $\mathbf{K}$, and values $\mathbf{V}$ by piping the input through three multi-layer perceptrons (MLPs). The general attention operation itself can be expressed as:
\begin{equation}
	Attention(\mathbf{Q},\mathbf{K},\mathbf{V})= softmax\left(\frac{\mathbf{Q}\mathbf{K}^T}{\tau}\right)\mathbf{V},
	\label{eq:attention}
\end{equation}
where the temperature $\tau$ scales the dot-product of $\textbf{Q}$ and $\textbf{K}^T$. There are two types of attention, self-attention and cross-attention. In self-attention, the queries originate from the same source as keys and values. While in cross-attention, queries do not share the same origin.

The DQ Perceiver combines both attention categories. In the cross-attention module, keys $\textbf{K}\in\mathbb{R}^{N\times d_k}$, and values $\textbf{V}\in\mathbb{R}^{N\times d_k}$ are projections of the input array (bag-of-instances) with shape $N\times C$. The queries $\mathbf{Q}_1\in\mathbb{R}^{M\times d_k}$ and $\mathbf{Q}_2\in\mathbb{R}^{1\times d_k}$ are projections of two learned latent arrays, one of size $M\times D$ and the other of shape $1\times D$, with $M+1\ll N$. As the query defines the shape of the output, the input array is distilled into a latent array of fixed size.

The dual-query module, illustrated in Figure \cref{fig:dq_perceiver_a}, leverages this behavior and creates two pathways. The first pathway is based on the regular Perceiver pipeline. Here we use $\mathbf{Q}_1$ to compress the input into a latent array, which afterward gets processed by the Latent Transformer. This module, shown in Figure \ref{fig:dq_perceiver_b}, performs self-attention on the latent array. The latent array is piped through the Latent Transformer $J$ times to improve the features. In the final step of this pathway, the latent array is averaged along the instance dimensions $M$ to obtain the self-attention token $t_{sa}$.

The second pathway is based on the idea of MIL-attention, where an attention-score $a$ weights each instance, so the aggregation function $g$ corresponds to weighted sum, see Equation~\ref{eq:weighted_sum}. This can be transferred to a single query cross-attention. Similar to the proposed method by \citet{Bergner2023}, the query  ($\mathbf{Q}_2$) of size $1\times d_k$ is used to predict attention scores for each projected instance $\mathbf{k}_i=\mathbf{W}_k \mathbf{h}_i$. Afterward, a second projected version of the instance $\mathbf{h_i}$, $\mathbf{v}_i=\mathbf{W}_v \mathbf{h}_i$ is scaled by the predicted attention score $a_i$ and summed up to build the MIL-attention token $t_{mil}$.
\begin{equation}
	t_{mil} = \sum_{i=1}^{N}a_i \mathbf{v}_i 
            = \sum_{i=1}^{N}a_i \mathbf{W}_v \mathbf{h}_i
            = \sum_{i=1}^{N} \frac{\text{exp}{\paren{s\paren{\mathbf{Q}_2,\mathbf{k}_i}}}}{\sum_{k=1}^{N} \text{exp}{\paren{s\paren{\mathbf{Q}_2,\mathbf{k}_k}}}} \mathbf{W}_v \mathbf{h}_i,
	\label{eq:weighted_sum}
\end{equation}
where $s\paren{\cdot,\cdot}$ denotes the scaled dot-product, given by $s\paren{\mathbf{Q}_2,\mathbf{k}}=\frac{\mathbf{Q}_2\mathbf{k}^T}{\tau}$ with temperature $\tau$ used as scaling factor. Each of the final bag representations $t_{sa}$ and $t_{mil}$ is processed in a separate MLP-based classifier in combination with a softmax operation to acquire the corresponding probability distribution $p_{sa}$ and $p_{mil}$, where $p_{sa}$ was determined by the self-attention based Perceiver branch and $p_{mil}$ predicted by the MIL pathway. The final output during inference is derived with a simple, balanced weighting mechanism, which can be expressed as $p= b p_{sa} + (1 - b) p_{mil}$, with $b$ as a hyper-parameter. This combination of outputs enables an architecture immanent supervision and the utilization of a self-distillation-based learning strategy. 

\subsubsection{Self-distillation Loss}
Self-distillation exploits components within an architecture to set up a knowledge-distillation-like learning scheme, in which shallow parts of a network are treated as an independent student architecture \cite{Zhang2019_SD, Zhang2022_SD}. For the DQ Perceiver, the final loss function $\mathcal{L}_{SD}$, is a combination of the main Cross-Entropy (CE) loss,  $\mathcal{L}_{CE}(p_{sa}, \fY)$ of the deepest part (Perceiver branch) and three additional self-distillation losses of the shallow part (MIL branch).

Like the main branch, the Cross-Attention Module also receives supervision by the bag label $\fY$. Moreover, the deeper Perceiver pathway supervises the Cross-Attention Module, using the Kullback-Leibler divergence between $p_{mil}$ and $p_{sa}$. This is complemented by an L2 loss, also called hint \cite{Romero2015}, inducing the MIL-attention token $t_{mil}$ to fit the self-attention token $t_{sa}$. Hyper-parameter $\alpha$ and $\lambda$ are used for balancing the contributions of the different loss terms. In our experiments, we empirically found the weighting factors $\alpha$ = 0.7 and $\lambda$ = 0.03 worked best for varying data sets.
\begin{equation}
	\mathcal{L}_{SD} = \mathcal{L}_{CE}(p_{sa}, \fY) + \alpha\mathcal{L}_{CE}(p_{mil}, \fY) + (1-\alpha)\mathcal{L}_{KL}(p_{mil}, p_{sa}) + \lambda\|t_{sa}-t_{mil}\|_2^2
	\label{eq:sd_loss}
\end{equation}

\section{Experiments and Results}
\label{sec:results}
\subsection{Experimental Design}
\label{subsec:design}
We thoroughly evaluate the DQ-MIL on three different histopathological datasets (Camelyon16, TCGA-BRCA, and TCGA-BLCA). The tasks are cancer classification and subtyping. Details regarding the different datasets, their curation, as well as about implementation are covered in the supplementary material. We report our evaluation using two metrics, area under the curve (AUC), and accuracy scores. Furthermore, we evaluated the localization of the most significant instances qualitatively. During pre-processing, each WSI is subdivided into patches, $x_i \in \mathbb{R}^{256\times256\times3}$ in magnification of $20\times$. Patches with background or artifacts are sorted out by combining threshold-based filtering \cite{Lu2020} with a pre-trained tissue segmentation U-Net \cite{Riasatian2020}. 

\subsection{Tumor Classification and Cancer Subtyping}
\label{subsec:tumor_class}
The two tasks we use for evaluation, tumor classification, and cancer subtyping, cover complementary challenges. Slides from the Camelyon16 dataset are highly unbalanced, where less than 10\% of the tissue area per slide covers positive instances (cancer) \cite{Li2020}. In contrast, The Genome Cancer Atlas (TCGA) datasets \cite{Liu2018}, which we use to test performance in cancer subtyping, require that the network does not just focus on small regions within the tissue but rather evaluate the global appearance of WSIs. For cancer subtyping, we use two publicly available datasets of different entities, breast cancer (BRCA) and bladder cancer (BLCA). The results of the classification are summarized in Table \ref{tbl:results_mil}. We realized that the pre-processing step, especially the Otsu-based filtering, has a strong impact on the final evaluation metrics. Thus, all values represented are based on experiments we run under the exact same conditions, using the dynamic meta-embedding approach for all of the different methods.
\begin{table}[!h]
	\centering
	\begin{tabular}{@{}lcccccc@{}}
		\hline
		\toprule
		\multirow{2}{*}{\shortstack[l]{Aggregation Method}}   &\multicolumn{2}{c}{Camelyon16}					&\multicolumn{2}{c}{TCGA-BRCA} 					&\multicolumn{2}{c}{TCGA-BLCA}		\\ 					\cmidrule(lr){2-3} \cmidrule(lr){4-5} \cmidrule(lr){6-7}		
							   & AUC 					& Accuracy				& AUC					& Accuracy				& AUC					& Accuracy			\\	\midrule		
		DS MIL \cite{Li2020}       & 0.8527        			& 0.8605   				& 0.9434			    & 0.8814			   	& 0.7312		    	& 0.8061			\\
		TransMIL \cite{Shao2021}   & 0.8559        			& 0.8450            	& 0.9308           		& 0.9040        		& 0.6769            	& \underline{0.8673}\\
		CLAM-SB \cite{Lu2020}	   & \underline{0.8946}     & \underline{0.8915}   	& \textbf{0.9455}		& \textbf{0.9266}		& \underline{0.7448} 	& 0.8061			\\	\midrule
		DQ-MIL-SD			   & \textbf{0.9594}		& \textbf{0.9457}  		& \underline{0.9441}    & \textbf{0.9266}   	& \textbf{0.8461}  		& \textbf{0.9184}		\\	\bottomrule
	\end{tabular}\\
	\hfill
	\caption{Performance evaluation of different MIL architectures on three medical datasets. The best performance is written in bold digits, and the second-best is underlined.}
	\label{tbl:results_mil}
\end{table}

Our proposed self-distilled DQ-MIL achieves state-of-the-art performance on the TCGA-BRCA dataset. For the Camelyon16 dataset, we achieve an improvement of up to 6.4\% in AUC and 5.4\% in accuracy. For the BLCA dataset, the improvement is even more significant, with up to 10.1\% in AUC and 5.1\% in accuracy compared to the second-best performing networks per metric.

\begin{figure}[h!]
	\centering
	\begin{subfigure}[b]{0.49\textwidth}
		\includegraphics[height=4.3cm]{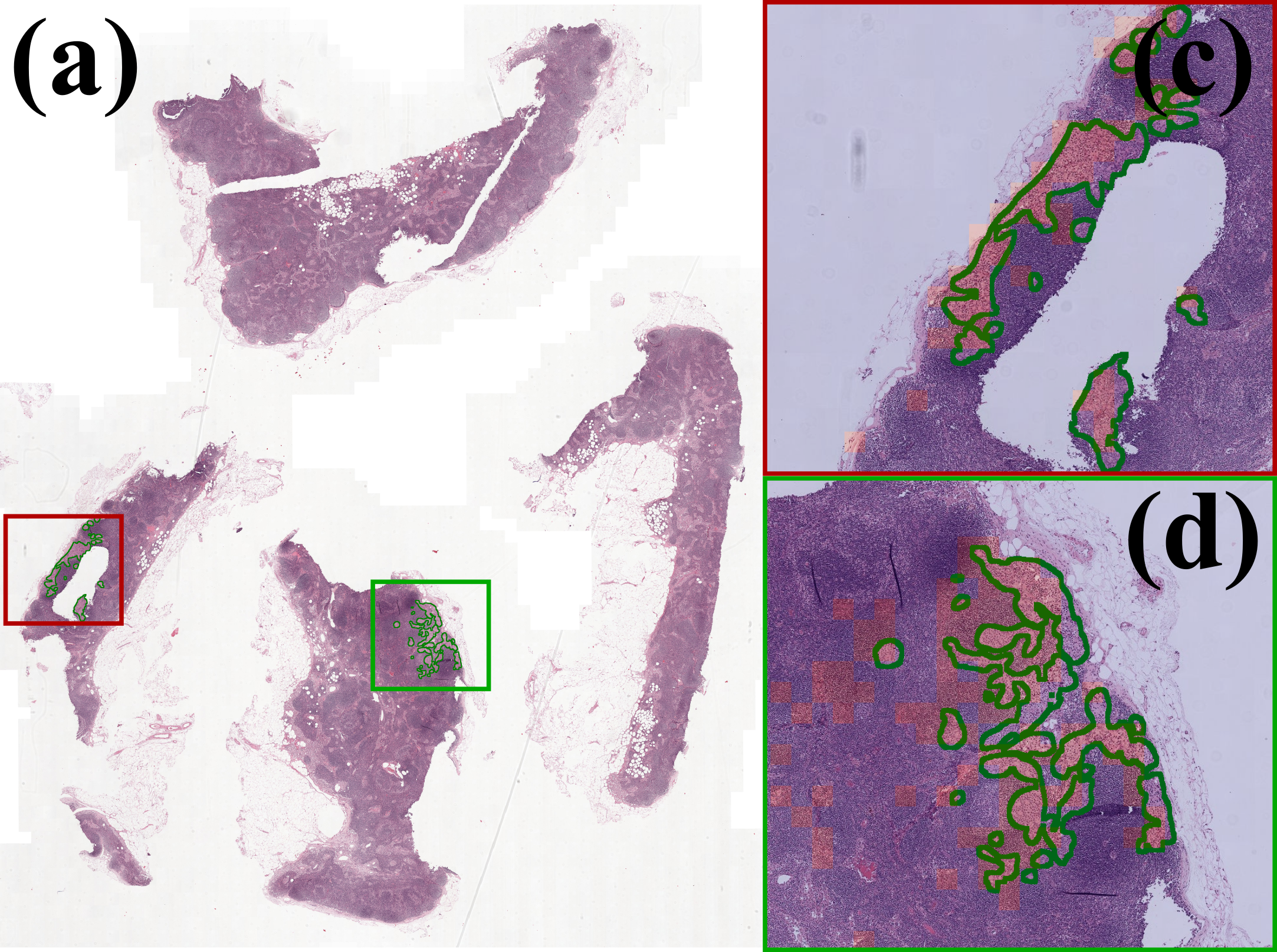}
	\end{subfigure}
	\begin{subfigure}[b]{0.49\textwidth}
		\includegraphics[height=4.3cm]{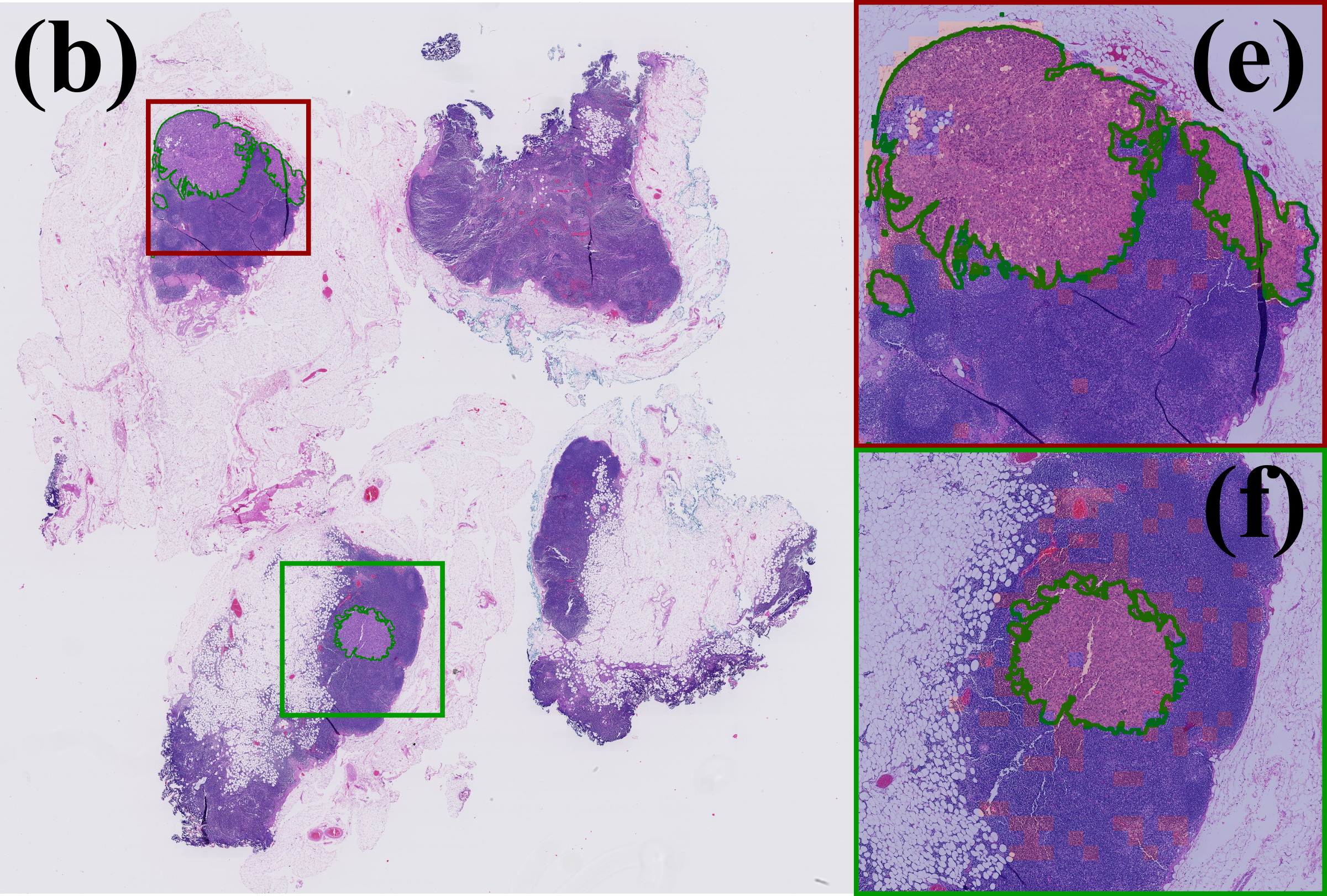}
	\end{subfigure}
	\caption{Visualization of the most significant attention scores. \textbf{(a)} and \textbf{(b)} show two WSIs from the Camelyon16 dataset. The red and green bounding boxes indicate the position of the cropped regions shown in \textbf{(c-f)}. The green contours in \textbf{(c-f)} indicate the cancerous regions annotated by pathologists. The attention scores are normalized per slide to [0,1], where the red colored regions in \textbf{(c-f)} highlight the patches with attention scores higher than 0.95.}
	\label{fig:qualtitative_results}
\end{figure}

To assess whether the DQ Perceiver is able to localize the most relevant areas with regard to the classification task, we also conduct a qualitative analysis, illustrated in Figure \ref{fig:qualtitative_results}. We utilize the pixel-wise annotations of the Camelyon16 dataset to evaluate the match between patches with top 5\% attention scores (highlighted in red Figure \ref{fig:qualtitative_results} \textbf{(c-f)}) and cancerous regions, annotated by domain experts (green contours in Figure \ref{fig:qualtitative_results} \textbf{(c-f)}). We can see that the DQ Perceiver is congruent with the annotated regions and is even able to detect small cancer areas, as shown in Figure \ref{fig:qualtitative_results} \textbf{(c)}.

\subsection{Ablation Study}
\label{subsec:ablation}
\subsubsection{Effects of the Individual Components of the Dual-Qyery Perceiver}
In this section, we evaluate how the individual components of our approach perform on the different classification tasks, the results are given in Table \ref{tbl:sq_vs_dq}.
Each sub-component, namely pure MIL Cross-Attention, the original Perceiver proposed by \citet{Jaegle2021}, and the Dual-Query Perceiver (DQ-MIL) without additional self-distillation is tested on all of the three medical datasets. For all sub-networks, we use a single cross-entropy loss during training. The final logits of the regular DQ-MIL are derived with the weighting mechanism, mentioned above, with $b=0.5$, leading to $p = \frac{1}{2}(p_{sa} + p_{mil})$. For this ablation study, we also use the dynamic meta-embedding strategy.
\begin{table}[h!]
	\centering
	\begin{tabular}{@{}lcccccc@{}}
		\hline
		\toprule
		\multirow{2}{*}{\shortstack[l]{Aggregation Method}}
		&\multicolumn{2}{c}{Camelyon16}				&\multicolumn{2}{c}{TCGA-BRCA} 					&\multicolumn{2}{c}{TCGA-BLCA}					\\	
		\cmidrule(lr){2-3} \cmidrule(lr){4-5} \cmidrule(lr){6-7}
		& AUC 				 	& Accuracy				& AUC					& Accuracy				& AUC					& Accuracy	  			\\	\midrule
		MIL Cross-Attention	 & \underline{0.9497}	& \underline{0.9380}	& 0.8940       			& \underline{0.9040}   	& 0.8172 				& \underline{0.8673}	\\	
		Perceiver   		 & 0.9439     		 	& 0.9147				& \textbf{0.9464}		& 0.8475   				& \textbf{0.8679}		& 0.8571				\\	
		DQ-MIL				 & 0.9099      		 	& 0.9147           		& 0.9362				& 0.8418				& 0.8303			 	& 0.8571				\\	\midrule
		DQ-MIL-SD		 & \textbf{0.9594}		& \textbf{0.9457} 		& \underline{0.9441}	& \textbf{0.9266}   	& \underline{0.8462}  	& \textbf{0.9184}		\\	\bottomrule \\
	\end{tabular}
 
	\caption{Comparison of the derivatives of the DQ-MIL-SD approach.}
	\label{tbl:sq_vs_dq}
\end{table}

We can see that the DQ-MIL-SD approach is slightly better or on par with its sub-components. The table also indicates that the self-distillation loss is the key element to boost the performance of the DQ-MIL-SD architecture and to join the benefits of the small MIL Cross-Attention model and the larger Perceiver. The table point out that for the cancer subtyping task, a correlation between the instances is slightly beneficial to improve on the AUC metric, whereas the MIL-attention approach achieves higher accuracy values. 

\subsubsection{Effect of the Dynamic Meta-Embedding Strategy}
We also conduct an ablation study to indicate the benefits and advantages of dynamic meta-embedding. Here we used the DQ-MIL-SD approach as our fixed evaluation model. We trained the model using the different embedding methods shown in Table \ref{tbl:meta_embedding}. Each embedding model varies in terms of architecture and SSL strategy. Furthermore, we compare the out-of-domain embedding methods ($^*$) with two in-domain pre-trained embedding methods ($^\dag$). The in-domain methods are a ResNet18 pre-trained on the Camelyon16 dataset using SimCLR \cite{Chen2020, Li2020} and a Vision Transformer (ViT) pre-trained on a large and comprehensive TCGA dataset covering multiple entities \cite{Chen_2022_CVPR}.

The performance evaluations show the advantage of the proposed Dynamic Meta-Embedder. It also indicates that the aggregation model can compensate for the embedding models' lack of domain knowledge. Although we were surprised to observe that the in-domain methods did not generalize well across our evaluation datasets, it resonates recent findings by \citet{Mcbee2023}.
\begin{table}[!h]
    \small
	\centering
	\begin{tabular}{@{}lcccccc@{}}
		\hline
		\toprule
		\multirow{2}{*}{\shortstack[l]{Embedding Method}}
								&\multicolumn{2}{c}{Camelyon16}				&\multicolumn{2}{c}{TCGA-BRCA} 					&\multicolumn{2}{c}{TCGA-BLCA}					\\	
								\cmidrule(lr){2-3} \cmidrule(lr){4-5} \cmidrule(lr){6-7}	
								&AUC 				 	& Accuracy			& AUC					& Accuracy				& AUC					& Accuracy	  			\\	\midrule
		ViT-B/8 Dino$^\dag$
		\cite{Chen_2022_CVPR}	& 0.7298     		 	& 0.7519 			& 0.8900        		& 0.8701        		& 0.7611      			& 0.8163				\\ 
		ResNet18 SimCLR$^\dag$
		\cite{Li2020}  			& 0.9136	 		 	& 0.9225           	& 0.8443         		& 0.8475          		& 0.7928				& 0.7857				\\ \midrule	
		ResNet50 SwAV$^*$		& \underline{0.9406}	& \underline{0.9302}& 0.9201				& 0.8927				& \underline{0.8045}  	& \underline{0.8776}	\\	
		ResNet50 DINO$^*$		& 0.8543	  		 	& 0.8837  			& 0.9347				& 0.8814				& 0.7747       			& 0.8265		 		\\	
		ViT-L/14 DINO v2$^*$	& 0.7474    		 	& 0.7984   			& \textbf{0.9704}   	& \textbf{0.9266}   	& 0.7405        		& 0.8367			\\ \midrule	
		Dynamic Meta-Embedder$^*$		& \textbf{0.9594}   	& \textbf{0.9457}  	& \underline{0.9441}    & \textbf{0.9266}   	& \textbf{0.8462}  		& \textbf{0.9184} 		\\	\bottomrule\\
	\end{tabular}
 
	\caption{Comparison of different embedding methods evaluated with a fixed DQ-MIL-SD aggregation model. The Dynamic Meta-Embedder utilizes all three SSL methods, pre-trained on ImageNet ($^*$). The ($^\dag$) indicates in-domain methods pre-trained on WSI patches.}
	\label{tbl:meta_embedding}
\end{table}

\section{Conclusion and Future Work}
In our work, we present a novel MIL approach called DQ-MIL-SD to the field of histopathological slide assessment. 
We introduce a dual-query cross-attention layer to combine single-token MIL-cross-attention with multi-token Perceiver cross- and self-attention in one architecture. By introducing a self-distillation loss, we can leverage the advantages of a small and a larger aggregation model. The proposed DQ Perceiver outperforms recent state-of-the-art approaches or is on par. In addition, combining multiple pre-trained embedders by the Dynamic Meta-Embedder ensures consistent performance across datasets. 
The next step will be to extend this approach to a multi-modal setting, allowing us to fully leverage the flexibility of the Perceiver and to explore its potential in the field of molecular subtyping.

\paragraph{Acknowledgements}
This work was sponsored by the Graduate School 2543/1 “Intraoperative Multisensory Tissue Differentiation in Oncology” (project ID 40947457), funded by the German Research Foundation (DFG - Deutsche Forschungsgemeinschaft). This work was also supported in part by the German Federal Ministry of Education and Research (BMBF): Tübingen AI Center, FKZ: 01IS18039A.
\bibliography{egbib}
\end{document}


\maketitle
\section{Datasets}
The experimental setup in this work utilizes three publicly available histopathological datasets: Camelyon16 \cite{Bejnordi2017}, The Cancer Genome Atlas (TCGA) Breast Invasive Carcinoma (BRCA) \cite{Thennavan2021}, and the TCGA Urothelial Bladder Carcinoma (BLCA) \cite{Robertson2017}. This section highlights the purposes of each dataset, the curation, and the pre-processing procedure. As mentioned in the main part of this work, all patches are extracted at 20$\times$ magnification in a non-overlapping manner with a size of 256$\times$256. 

\subsection{Camelyon16}
\label{sec:camelyon}
The Camelyon16 dataset \cite{Bejnordi2017} consists of 399 hematoxylin and eosin (H\&E) stained lymph node sections, scanned and stored as whole-slide images (WSIs). Each slide is fully annotated and permits pixel-wise detection of breast cancer metastasis. We focus on slide-level cancer classification in our weakly supervised setup and ignore the pixel-wise annotations. The WSIs are labeled as "tumor" as soon as they incorporate annotated cancerous regions, otherwise they are "normal". We follow the official dataset split with 270 training samples (110 tumor, 160 normal) and 129 test samples (49 tumor, 80 normal). During pre-processing, we combine threshold-based filtering \cite{Lu2020} with a pre-trained U-Net \cite{Riasatian2020} for tissue segmentation, yielding about 11,500 patches per slide.

\subsection{TCGA-BRCA}
\label{sec:brca}
The TCGA-BRCA \cite{Thennavan2021} contains 1,133 diagnostics digital H\&E slides of invasive breast cancer and is made available by the National Cancer Institute (NCI) Genomic Data Commons (GDC) \cite{Grossman2016}. The dataset covers 15 histological types and can be augmented with additional modalities such as genomic data. Following the experimental design of \citet{Chen_2022_CVPR}, we focus on classifying the two most frequent histological types of breast cancer: invasive ductal carcinoma (IDC) and invasive lobular carcinoma (ILC). We apply a stratified data split with a ratio of 80:20 (training:test) on the patient-level, which leads to 698 training samples (578 IDC, 120 ILC) and 177 test samples (148 IDC, 29 ILC). As the WSIs do not contain a 20$\times$ magnification, we extract patches of size 512 at magnification 40$\times$ and apply a downsampling operation of factor 2 to acquire patches of size 256$\times$256. The remaining steps during pre-processing are the same as described in Section \ref{sec:camelyon}, leading to roughly 11,000 patches per slide.

\subsection{TCGA-BLCA}
The TCGA-BLCA dataset \cite{Robertson2017} is also published by the NCI GDC \cite{Grossman2016} and comprises 449 labeled diagnostic H\&E WSIs of muscle-invasive bladder cancer (MIBC). In our experiments, we intend to classify the slides into two histological types: papillary MIBC and non-papillary MIBC. We exploit the same procedure as in Section \ref{sec:brca} and apply a patient-level data split with 80\% training cases (351 WSIs) and 20\% test cases (98 WSIs). After pre-processing, we acquire approximately 16,500 patches per slide.

\section{Implementation Details}
To train the DQ-MIL architecture, a self-distillation loss $\mathcal{L}_{SD}$, inspired by \citet{Zhang2019_SD, Zhang2022_SD}, is utilized and combined with a Lookahead RAdam optimizer \cite{Zhang2019, Liu2019}. For all experiments, a learning rate of $2\times10^{-4}$ and a weight decay of $10^{-5}$ is used \cite{Shao2021}. The mini-batch during training is set to one bag-of-instances (1 WSI). Following \citet{Jaegle2021}, a truncated normal distribution with $\mu=0$, $\sigma=0.02$, and truncation bounds of [-2, 2] is used to randomly initialize the latent representations ($\mathbf{Q_1}$,  $\mathbf{Q_2}$). The hyper-parameter setting of the DQ-MIL architecture used for the experiments, results in a computational complexity of 25 GFLOPS, which is decreased compared to TransMil \cite{Shao2021} with 40 GFLOPS and DS MIL \cite{Li2020} with 45 GFLOPS. 

\section{Ablation Study}
\subsection{Temperature-Based Instance Masking}
Motivated by the results of the Iterative Patch Selection (IPS) module \cite{Bergner2023}, which condenses a bag into its M most salient instances, we conduct an ablation study to explore the potential of temperature $\tau$ for implicit instance masking. As shown in the main section of this work, the general attention operation, based on queries $\mathbf{Q}$, keys $\mathbf{K}$, values $\mathbf{V}$, and temperature $\tau$, can be expressed as:
\begin{equation}
	Attention(\mathbf{Q},\mathbf{K},\mathbf{V})= softmax\left(\frac{\mathbf{Q}\mathbf{K}^T}{\tau}\right)\mathbf{V}.
	\label{eq:attention}
\end{equation}
In standard self-attention, $\tau$ serves to decouple the attention scores from the inner channel dimension $d_k$. Therefore, parameter $\tau$ is given by $\tau=\sqrt{d_k}$.
In contrast to \citet{Bergner2023}, our idea is not to reduce the computational burden. We aim to sharpen the training signal by implicitly masking out less significant instances. Therefore, we decrease the temperature $\tau$ to collapse the probability distribution to the most essential instances. To explore the effect of this approach, we conducted experiments with various values for $\tau$. The results are shown in Table \ref{tbl:sharpeing}.
\begin{table}[!h]
	\centering
	\begin{tabular}{@{}lcccccc@{}}
		\hline
		\toprule
		\multirow{2}{*}{\shortstack[l]{Temperature}}
		&\multicolumn{2}{c}{Camelyon16}				&\multicolumn{2}{c}{TCGA-BRCA} 					&\multicolumn{2}{c}{TCGA-BLCA}					\\	
		\cmidrule(lr){2-3} \cmidrule(lr){4-5} \cmidrule(lr){6-7}	
								&AUC 				& Accuracy			 & AUC				& Accuracy				& AUC				& Accuracy			\\	\midrule
		$\tau = \sqrt{d_k} = 8$	& \underline{0.9594} & \textbf{0.9457}   & \textbf{0.9441}   & \textbf{0.9266}  	& \textbf{0.8462}	& \textbf{0.9184}		\\	
		$\tau = 1$			 	& 0.9487		 	 & \textbf{0.9457}	 & \underline{0.9369}&\underline{0.9039}	& \underline{0.8452}& 0.8061			\\	
		$\tau = 1/8$			& 0.9556			 & 0.9380			 & 0.9306 			 & 0.8249 				& 0.8081			& 0.7959 			\\	
		$\tau = 1/16$  			& \textbf{0.9651} 	 & \textbf{0.9457}	 & 0.9359         	 & 0.8531          		& 0.8027			& \textbf{0.9184}\\	\bottomrule
	\end{tabular}\\
	\hfill
	\caption{Comparison of different temperature values, evaluated with a fixed DQ-MIL-SD aggregation model.}
	\label{tbl:sharpeing}
\end{table}\\
Although we achieve an improvement of the AUC metric on Camelyon16, which resonates with the insights from \citet{Bergner2023}, the potential of implicit instance selection using temperature $\tau$ is limited. Collapsing the probability distributions by decreasing $\tau$ seems only beneficial for unbalanced bags-of-instances, given in the Camelyon16 dataset. For other tasks, such as histological subtyping, temperature-based instance masking may even be detrimental to the overall performance. An alternative approach could be to convert the hyperparameter $\tau$ into a trainable parameter \cite{Guo2017}.
\bibliography{egbib}